\documentclass[11pt]{article} 
\usepackage{etoolbox}
\makeatletter
\patchcmd{\maketitle}
 {\def\@makefnmark}
 {\def\@makefnmark{}\def\useless@macro}
 {}{}
\makeatother
\usepackage{geometry}
\usepackage{lmodern}
\usepackage{hyperref}       
\hypersetup{
    colorlinks,
    breaklinks,
    linkcolor = blue,
    citecolor = blue,
    urlcolor  = black,
} 
\usepackage{natbib}
\usepackage{comment}
\usepackage[utf8]{inputenc} 

\usepackage[T1]{fontenc}    
\usepackage{hyperref}       
\usepackage{url}            
\usepackage{booktabs}       
\usepackage{amsfonts}       
\usepackage{nicefrac}       
\usepackage{microtype}      
\usepackage{xcolor}         
\usepackage{amsmath}
\usepackage{amssymb}
\usepackage{mathtools}
\usepackage{amsthm}
\usepackage{algorithm}
\usepackage{algorithmic}
\usepackage{arydshln}
\usepackage{bigstrut}
\usepackage{bm}
\usepackage{booktabs}
\usepackage{color}
\usepackage[english]{babel}
\usepackage{enumitem}
\usepackage{listings}
\usepackage{multirow}
\usepackage{multicol}
\usepackage{newtxmath}
\usepackage{rotating}
\usepackage{stmaryrd}
\usepackage{subfigure}
\usepackage{tablefootnote}
\usepackage{longtable}
\usepackage{threeparttable}
\usepackage{amssymb}
\usepackage{pifont}
\usepackage{wrapfig}

\usepackage{makecell}

\theoremstyle{plain}
\newtheorem{theorem}{Theorem}
\newtheorem{proposition}{Proposition}
\newtheorem{lemma}{Lemma}

\theoremstyle{definition}
\newtheorem{definition}{Definition}
\newtheorem{assumption}{Assumption}
\theoremstyle{remark}

\title{Recent advances in weakly supervised learning: New supervision paradigms, assumption relaxations, and practical solutions}
\author{Wei Wang$^{1,2}$, Gang Niu$^{1}$, Masashi Sugiyama$^{1,2}$ \\
  $^1$ RIKEN, Tokyo, Japan\\
  $^2$ The University of Tokyo, Chiba, Japan\\
  \texttt{wwangwitsel@gmail.com}
}
\date{}
\begin{document}
\maketitle
\begin{abstract}
Deep learning has achieved great success in recent years thanks to the availability of high-quality, well-annotated training data. However, this requirement is often not met in real-world applications. Weakly supervised learning aims to train an accurate model with incomplete, inexact, or inaccurate supervision. In this chapter, we will discuss recent advances in this field, including new supervision paradigms, relaxed assumptions, and practical solutions. First, we introduce a new weakly supervised binary classification problem called confidence-difference classification and propose consistent approaches to solve it. Next, we investigate complementary-label learning, a weakly supervised multi-class classification problem. Our proposed approaches are based on more relaxed assumptions about the data generation process than existing consistent approaches. Lastly, we present an evaluation framework for partial-label learning, another popular multi-class weakly supervised learning problem, in order to promote fair and realistic evaluation of algorithms in this field.
\end{abstract}
\section{Introduction}
In recent years, deep learning has achieved remarkable success in various real-world applications thanks to its strong generalization capabilities. However, this success depends heavily on the availability of large-scale, accurately labeled training data, which is often costly or impractical to obtain in real-world scenarios. Weakly supervised learning offers a promising solution, allowing models to be trained with imperfect supervision while maintaining performance comparable to that of fully supervised approaches~\citep{sugiyama2022machine}. 

Weakly supervised learning~(WSL) lies at the intersection of supervised and unsupervised learning. In this chapter, we will focus primarily on weakly supervised classification, which involves constructing a binary or multi-class classifier that assigns discrete labels to test data using only weakly supervised data. According to the types of weak supervision, WSL can be roughly categorized into learning with incomplete, inexact, or inaccurate supervision~\citep{zhou2018brief}. This chapter introduces several recent advances in weakly supervised learning, including new supervision paradigms, assumption relaxations, and practical solutions.

\begin{figure}[t]
  \centering
  \includegraphics[width=0.8\linewidth]{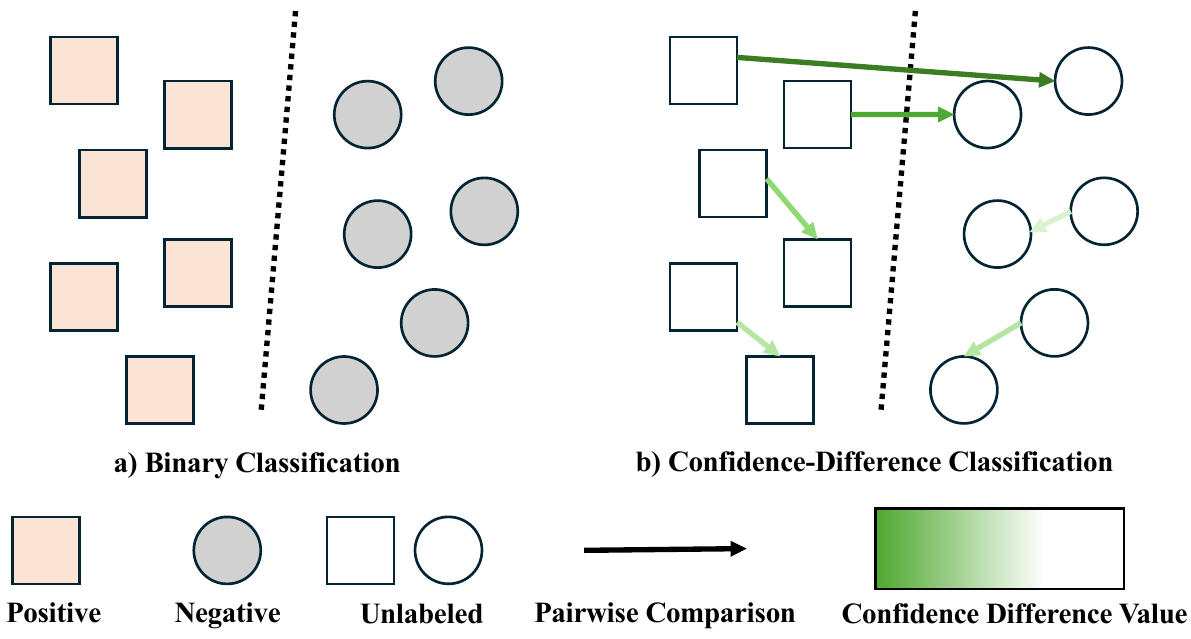}
  \caption{Comparison of ordinary binary classification and confidence-difference classification investigated in Section~\ref{sec:confdiff}. In confidence-difference classification, we are given unlabeled data pairs with confidence difference. The goal is the same as that in binary classification. Here, the color of the directional line indicates the value of the confidence difference.}
  \label{fig:confdiff}
\end{figure}

Section~\ref{sec:confdiff} investigates a novel weakly supervised binary classification framework called \emph{confidence-difference classification}, which weakens the supervision requirements when leveraging soft-label information. While soft-label learning has demonstrated superior performance over hard-label approaches in various domains~\citep{szegedy2016rethinking,yuan2023learning}, the acquisition of pointwise confidence labels for all training examples remains impractical in real-world applications~\citep{collins2022eliciting,shinoda2020binary,sucholutsky2023on}. To address this limitation, the confidence-difference paradigm operates on unlabeled data pairs annotated only with relative confidence differences indicating the probability difference of being positive rather than absolute confidence values~(see Figure~\ref{fig:confdiff}). We introduce unbiased~(see Section~\ref{sec:ure_confdiff}) and corrected risk estimators~(see Section~\ref{sec:confdiff_cre}) for this problem, which have both consistency guarantees and convergence properties. In addition, we provide theoretical analysis that quantifies how estimation errors in confidence values and class priors affect the final classifier performance~(see Section~\ref{sec:confdiff_robustness}). 

\begin{figure}[t]
  \centering
  \includegraphics[width=0.8\linewidth]{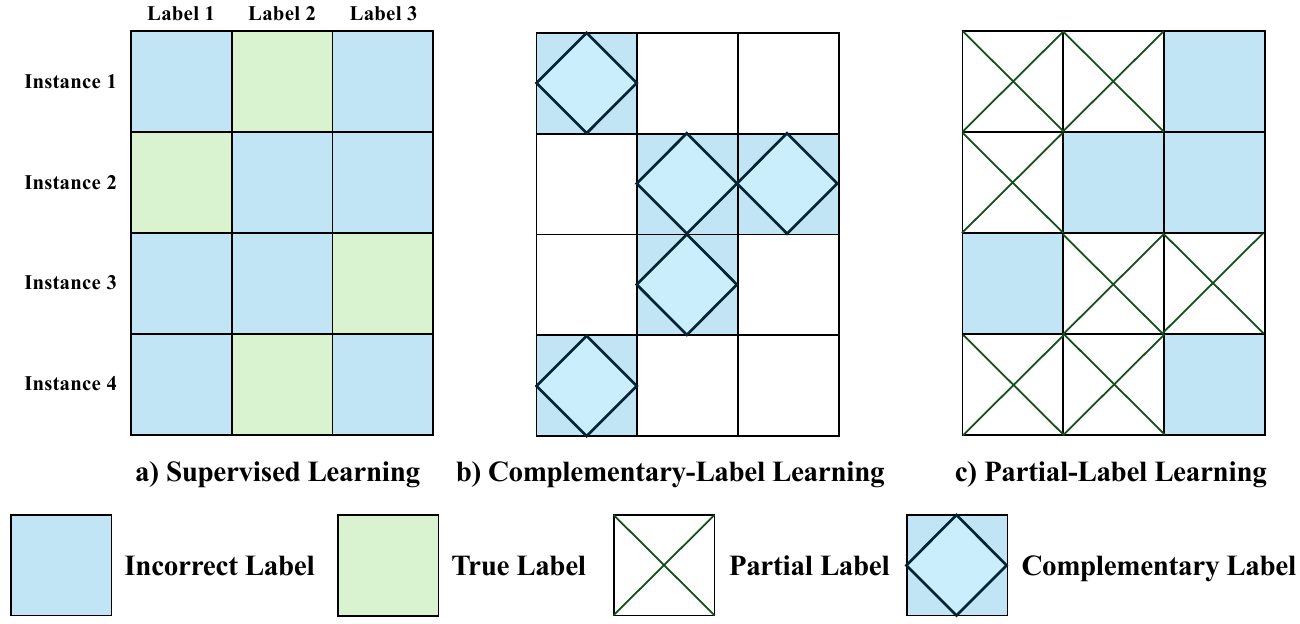}
  \caption{Comparison of ordinary multi-class classification, complementary-label learning investigated in Section~\ref{sec:scarce}, and partial-label learning investigated in Section~\ref{sec:plench}. Complementary-label learning and partial-label learning are two types of weakly supervised, multi-class classification problems involving inexact supervision.}
  \label{fig:plcl}
\end{figure}

\begin{sloppypar}
Section~\ref{sec:scarce} introduces a novel consistent learning approach for complementary-label~(CL) learning which attempts to relax the restrictive assumptions for CL learning~\citep{wang2024learning}. CL learning is a multi-class weakly supervised learning problem where training examples are annotated only with labels to which they do not belong~(see Figure~\ref{fig:plcl})~\citep{ishida2017learning}. While existing consistent methods either require restrictive uniform distribution assumptions~\citep{ishida2019complementary,feng2020learning} or depend on ordinary-label data to estimate transition matrices~\citep{yu2018learning}, such conditions often fail in practical settings~\citep{wang2025climage}. Our work breaks this dependency by deriving an unbiased risk estimator based on a more practical data generation assumption~(see Section~\ref{sec:scarce_ure}). Then, we propose a risk-correction approach to combat overfitting problems when using deep models~(see Section~\ref{sec:scarce_cre}). The consistency and convergence rate of the estimation error are also established.  
\end{sloppypar}

Section~\ref{sec:plench} presents the first comprehensive benchmarking framework for the realistic evaluation of partial-label~(PL) learning~\citep{wang2025realistic}. In PL learning, each training example is associated with multiple candidate labels, one of which is the ground truth~(see Figure~\ref{fig:plcl})~\citep{cour2011learning,lv2020progressive,feng2020provably}. Despite significant algorithmic advances in PL learning in recent years, our analysis reveals critical gaps in current evaluation practices: First, there is an absence of principled model selection methods, despite their substantial impact on performance~\citep{wang2025realistic}. Second, there are inconsistent experimental protocols that compromise fair comparisons. These shortcomings have led to a systematic underestimation of earlier, simpler algorithms relative to more complex, recent approaches~\citep{lv2020progressive,feng2020provably}. We address these limitations by providing theoretically grounded model selection criteria designed specifically for PL learning and PLENCH~(PL learning bENCHmark), a standardized evaluation protocol~(see Section~\ref{model_selection_section}). PLENCH provides a reproducible and fair comparison of PL learning algorithms under realistic conditions and serves as an essential testbed for future research in this field~(see Section~\ref{sec:plench_settings}). We anticipate that PLENCH will lay the groundwork for the more rigorous development and evaluation of PL learning methods. 

In summary, this chapter presents several recent advances in weakly supervised learning, including new supervision paradigms~(confidence difference classification), assumption relaxations~(selected-completely-at-random CL learning), and practical solutions~(the first PL learning benchmark). 
\section{Preliminaries}
\label{sec:preliminary}
In this section, we introduce the basic notation and background of binary and multi-class classification, as well as complementary-label and partial-label learning.
\subsection{Notations}
Let $\mathcal{X} = \mathbb{R}^d$ denote the $d$-dimensional feature space and $\mathcal{Y}$ denote the label space. For binary classification, we have $\mathcal{Y}=\left\{+1, -1\right\}$. For multi-class classification, we have $\mathcal{Y}=\left\{1,2,\ldots,q\right\}$. Let $p(\bm{x}, y)$ denote the unknown joint probability density over the random variables $(\bm{x}, y)\in \mathcal{X}\times \mathcal{Y}$. An example is denoted by $(\bm{x}, y)$. Here, $\bm{x}\in\mathcal{X}$ denotes an instance that characterizes the properties of this example. Besides, $y\in\mathcal{Y}$ denotes the label of this example, which encodes the corresponding semantic information. Let $p(\bm{x})$ denote the marginal density of the instances. For binary classification, let $\pi_{+}=p(y=+1)$ and $\pi_{-}=p(y=-1)$ denote the class prior probabilities for the positive and negative classes respectively. Furthermore, let $p_{+}(\bm{x})=p(\bm{x}|y=+1)$ and $p_{-}(\bm{x})=p(\bm{x}|y=-1)$ denote the class-conditional probability densities of positive and negative data respectively. For multi-class classification, we use $\pi_{k}=p(y=k)$ to denote the class prior probability for the $k$-th class. The goal of binary or multi-class classification is to learn a mapping from the feature space $\mathcal{X}$ to the label space $\mathcal{Y}$. For binary classification, let $g:\mathcal{X}\rightarrow \mathbb{R}$ denote the binary classifier to learn. Then, we use $\ell\left(g(\bm{x}), y\right)$ to denote a binary loss function that characterizes the differences between the model output $g(\bm{x})$ and the ground-truth $y$ for $x$. For multi-class classification, let $\bm{f}:\mathcal{X}\rightarrow\mathbb{R}^{q}$ denote the multi-class classifier to be learned. Here, $\bm{f}(\bm{x})=\left[f_{1}(\bm{x}),f_{2}(\bm{x}),\ldots,f_{q}(\bm{x})\right]$ is a $q$-dimensional vector which denotes the model output. We use $f:\mathcal{X}\rightarrow\mathcal{Y}$ to denote the label predicted by the model, which corresponds to the dimension of the maximum model output, i.e., 
\begin{equation}    
f(\bm{x}) = \mathop{\arg\max}_{k\in\mathcal{Y}}~f_{k}(\bm{x}).
\end{equation}
Then, we use $\mathcal{L}\left(\bm{f}(\bm{x}), y\right)$ to denote a multi-class loss function which characterizes the difference between the model output $\bm{f}(\bm{x})$ and the ground-truth label $y$ for $x$.
\subsection{Ordinary Binary Classification}
In ordinary binary~(OB) classification, our goal is to learn a binary classifier $g(\bm{x})$ which minimizes the following test error for binary classification:
\begin{equation}
R_{0\mathrm{-}1}^{\rm OB}(g)=\mathbb{E}_{p(\bm{x},y)}\mathbb{I}(yg(\bm{x}) < 0),
\end{equation}
where $\mathbb{I}$ is an indicator function that returns $1$ if the predicate holds; otherwise it returns $0$. However, the 0-1 loss function is non-smooth and difficult to optimize. Therefore, we replace the 0-1 loss function with a classification-calibrated loss function $\ell$, and the classification risk for binary classification is defined as 
\begin{equation}\label{eq:binary_class_risk}
R^{\rm OB}(g) = \mathbb{E}_{p(\bm{x},y)}[\ell\left(g(\bm{x}),y\right)].
\end{equation}
For binary classification, a margin loss $\ell^{\mathrm M}$ is often adopted to instantiate the loss function, such as the logistic loss. The classification risk can also be expressed as
\begin{equation}
R^{\rm OB}(g) = \mathbb{E}_{p(\bm{x},y)}[\ell^{\mathrm M}\left(yg(\bm{x})\right)].
\end{equation}

We cannot compute the expectation in Eq.~(\ref{eq:binary_class_risk}) directly.
To cope with this problem, we typically assume that we are given a training set i.i.d.~sampled from $p\left(\boldsymbol{x},y\right)$:
\begin{equation}
\mathcal{D}^{\mathrm{OB}} =\left\{\left(\boldsymbol{x}^{\mathrm{OB}}_{i},y^{\mathrm{OB}}_{i}\right)\right\}_{i=1}^{n^{\mathrm{OB}}} \stackrel{\text { i.i.d. }}{\sim} p\left(\boldsymbol{x},y\right).
\end{equation}
Then, we conduct \emph{empirical risk minimization} by minimizing the following empirical risk:
\begin{equation}
\widehat{R}^{\mathrm{OB}}(g)=\frac{1}{n^{\mathrm{OB}}}\sum\nolimits_{i=1}^{n^{\mathrm{OB}}}\ell\left(g\left(\boldsymbol{x}^{\mathrm{OB}}_{i}\right),y^{\mathrm{OB}}_{i}\right).
\end{equation}

\subsection{Ordinary Multi-Class Classification}
In ordinary multi-class~(OM) classification, our goal is to learn a multi-class classifier $f(\bm{x})$ which minimizes the following test error for multi-class classification:
\begin{equation}
R_{0\mathrm{-}1}^{\rm OM}(f)=\mathbb{E}_{p(\bm{x},y)}\mathbb{I}(f(\bm{x})\neq y).
\end{equation}
Similar to binary classification, we replace the 0-1 loss function with a classification-calibrated loss function $\mathcal{L}$, and the classification risk for multi-class classification is defined as 
\begin{equation}\label{eq:multi_class_risk}
R^{\rm OM}(g) = \mathbb{E}_{p(\bm{x},y)}[\mathcal{L}\left(f(\bm{x}),y\right)].
\end{equation}
We cannot compute the expectation in Eq.~(\ref{eq:multi_class_risk}) directly. To cope with this problem, we typically assume that we are given a training set i.i.d.~sampled from $p\left(\boldsymbol{x},y\right)$:
\begin{equation}
\mathcal{D}^{\mathrm{OM}} =\left\{\left(\boldsymbol{x}^{\mathrm{OM}}_{i},y^{\mathrm{OM}}_{i}\right)\right\}_{i=1}^{n^{\mathrm{OM}}} \stackrel{\text { i.i.d. }}{\sim} p\left(\boldsymbol{x},y\right).
\end{equation}
Then, we conduct \emph{empirical risk minimization} by minimizing the following empirical risk for multi-class classification:
\begin{equation}
\widehat{R}^{\mathrm{OM}}(g)=\frac{1}{n^{\mathrm{OM}}}\sum\nolimits_{i=1}^{n^{\mathrm{OM}}}\mathcal{L}\left(f\left(\boldsymbol{x}^{\mathrm{OM}}_{i}\right),y^{\mathrm{OM}}_{i}\right).
\end{equation}
\subsection{Complementary-Label Learning}
In CL learning, each training example is associated with one or multiple complementary labels specifying the classes to which the example does not belong. Let $\mathcal{D}^{\mathrm{CL}}=\left\{\left(\bm{x}^{\mathrm{CL}}_i, \bar{Y}^{\mathrm{CL}}_i\right)\right\}_{i=1}^{n^{\mathrm{CL}}}$ denote a CL training set sampled i.i.d.~from an unknown density $p(\bm{x}, \bar{Y})$. Here, $\bm{x} \in \mathcal{X}$ is a feature vector, and $\bar{Y} \subseteq \mathcal{Y}$ is a CL set associated with $\bm{x}$. In the literature, CL learning can be categorized into single CL learning when we have $\bar{Y}=\left\{\bar{y}\right\}$ and $|\bar{Y}|=1$~\citep{ishida2017learning,
gao2021discriminative,liu2023consistent}, and multiple CL learning when we have $1\leq|\bar{Y}|\leq q-1$~\citep{feng2020learning}. The task of CL learning is to learn a multi-class classifier $f:\mathcal{X}\rightarrow \mathcal{Y}$ from $\mathcal{D}^{\mathrm{CL}}$. 
\subsection{Partial-Label Learning}
Let $\left(\bm{x},S\right)$ denote a PL example where $\bm{x} \in \mathcal{X}$ is a feature vector and $S \subseteq \mathcal{Y}$ is a candidate label set associated with $\bm{x}$. The basic assumption of PLL is that the ground-truth label $y$ of $\bm{x}$ is concealed within its candidate label set $S$, i.e., $y\in S$. The task of PLL is to learn a multi-class classifier $f$ from a PL training set $\mathcal{D}^{\rm Tr}=\{\left(\bm{x}^{\rm Tr}_i,S^{\rm Tr}_i\right)\}_{i=1}^{n^{\rm Tr}}$. 
\section{ConfDiff Classification}
\label{sec:confdiff}
Recently, learning with \emph{soft labels} has been shown to achieve better performance than learning with \emph{hard labels} in terms of model generalization, calibration, and robustness. However, collecting pointwise labeling confidence for all training examples can be challenging and time-consuming in real-world scenarios. This section delves into a novel weakly supervised binary classification problem called \emph{confidence-difference (ConfDiff) classification}~\citep{wang2023binary}. Instead of pointwise labeling confidence, we are given only unlabeled data pairs with confidence difference that specifies the difference in the probabilities of being positive. We propose a risk-consistent approach to tackle this problem and show that the estimation error bound achieves the optimal convergence rate. We also introduce a risk-correction approach to mitigate overfitting problems, whose consistency and convergence rate are also proven.
\subsection{Motivation}
Learning with soft labels has been shown to achieve better performance than learning with hard labels in the context of supervised learning~\citep{yuan2023learning}, where each example is equipped with \emph{pointwise labeling confidence} indicating the degree to which the labels describe the example. The advantages have been validated in many aspects, including model generalization~\citep{yuan2020revisiting,ishida2023is}, calibration~\citep{muller2019does,wang2021rethinking}, and robustness~\citep{lukasik2020does,pang2020bag}. For example, with the help of soft labels, knowledge distillation~\citep{yuan2020revisiting,hinton2015distilling} transfers knowledge from a large teacher network to a small student network. The student network can be trained more efficiently and reliably with the soft labels generated by the teacher network~\citep{phuong2019towards,park2019relational,gou2021knowledge}.

However, collecting a large number of training examples with pointwise labeling confidence may be demanding under many circumstances since it is challenging to describe the labeling confidence for each training example exactly~\citep{collins2022eliciting,shinoda2020binary,sucholutsky2023on}. Different annotators may give different values of pointwise labeling confidence to the same example due to personal biases, and it has been demonstrated that skewed confidence values can harm classification performance~\citep{shinoda2020binary}. Besides, giving pointwise labeling information to large-scale datasets is also expensive, laborious, and even unrealistic in many real-world scenarios~\citep{wei2022learning,karimi2020deep,ratner2016data}.
On the contrary, leveraging supervision information of pairwise comparisons may ameliorate the biases of skewed pointwise labeling confidence and save labeling costs. Following this idea, we investigate a more practical problem setting for binary classification in this section, where we are given \emph{unlabeled data pairs with confidence difference} indicating the difference in the probabilities of being positive. Collecting confidence difference for training examples in pairs is much cheaper and more accessible than collecting pointwise labeling confidence for all the training examples. 

Take click-through rate prediction in recommender systems~\citep{zhang2019deep,jiang2022adaptive} for example. The combinations of users and their favorite/disliked items can be regarded as positive/negative data. Collecting training data takes work to distinguish between positive and negative data.
Furthermore, the pointwise labeling confidence of training data may be difficult to be determined due to the extraordinarily sparse and class-imbalance problems~\citep{yao2021self}. Therefore, the collected confidence values may be biased. However, collecting the difference in the preference between a pair of candidate items for a given user is more accessible and may alleviate the biases. In \citet{wang2023binary}, we provided an experiment on recommender systems to illustrate it.

Take the disease risk estimation problem for another example. Given a person's attributes, the goal is to predict the risk of having some disease. When asking doctors to annotate the probabilities of having the disease for patients, it takes work to determine the exact values of the probabilities. Furthermore, the probability values given by different doctors may differ due to their diverse backgrounds. On the other hand, it is much easier and less biased to estimate the relative difference in the probabilities of having the disease between two patients. Therefore, the problem of learning with confidence difference is of practical research value, but has yet to be investigated in the literature.  

\subsection{Problem Definition}
In this subsection, the formal definition of confidence difference is given firstly. Then, we elaborate the data generation process of ConfDiff data.
\begin{definition}[Confidence Difference]\label{df1}
The confidence difference $c\left(\bm{x}, \bm{x}'\right)$ between an unlabeled data pair $\left(\bm{x}, \bm{x}'\right)$ is defined as 
\begin{equation}
c\left(\bm{x}, \bm{x}'\right) 
= p\left(y'=+1|\bm{x}'\right) - p\left(y=+1|\bm{x}\right).
\end{equation}
\end{definition}
As shown in the definition above, the confidence difference denotes the difference in the class posterior probabilities between the unlabeled data pair, which can measure how confident the pairwise comparison is. In ConfDiff classification, we are only given $n$ unlabeled data pairs with confidence difference $\mathcal{D}^{\mathrm{CD}}=\left\{\left(\left(\bm{x}_{i}^{\mathrm{CD}}, \bm{x}_{i}^{\mathrm{CD}'}\right), c_{i}\right)\right\}_{i=1}^{n}$. Here, $c_{i}=c\left(\bm{x}_{i}^{\mathrm{CD}}, \bm{x}_{i}^{\mathrm{CD}'}\right)$ is the confidence difference for the unlabeled data pair $\left(\bm{x}_{i}^{\mathrm{CD}}, \bm{x}_{i}^{\mathrm{CD}'}\right)$. Furthermore, the unlabeled data pair $\left(\bm{x}_{i}^{\mathrm{CD}}, \bm{x}_{i}^{\mathrm{CD}'}\right)$ is assumed to be drawn from a probability density $p\left(\bm{x}, \bm{x}'\right)=p\left(\bm{x}\right)p\left(\bm{x}'\right)$. This indicates that $\bm{x}_{i}^{\mathrm{CD}}$ and $\bm{x}_{i}^{\mathrm{CD}'}$ are two i.i.d.~instances sampled from $p(\bm{x})$. It is worth noting that the confidence difference $c_{i}$ will be positive if the second instance $\bm{x}_{i}^{\mathrm{CD}'}$ has a higher probability to be positive than the first instance $\bm{x}_{i}^{\mathrm{CD}}$, and will be negative otherwise. During the data collection process, the labeler can first sample two unlabeled data independently from the marginal distribution $p(\bm{x})$, then provide the confidence difference for them.
\subsection{Methodology}\label{section_algo}
In this section, we introduce our proposed approaches with theoretical guarantees. Besides, we show the influence of an inaccurate class prior probability and noisy confidence difference theoretically. Furthermore, we introduce a risk-correction approach to improve the generalization performance. Due to page limitations, detailed proofs of theorems and lemmas, as well as experimental results, can be found in \citet{wang2023binary}.
\subsubsection{Unbiased Risk Estimator}\label{sec:ure_confdiff}
In this subsection, we show that the classification risk in Eq.~(\ref{eq:binary_class_risk}) can be expressed with ConfDiff data in an equivalent way. We rewrite the classification risk as the expectation w.r.t.~the densities of unlabeled data pairs of a loss function containing confidence difference. The detailed proof can be found in \citet{wang2023binary}.
\begin{theorem}\label{ccrisk}
The classification risk in Eq.~(\ref{eq:binary_class_risk}) can be equivalently expressed as 
\begin{equation}\label{ccrisk_eqn}
R_{\rm CD}(g) = \mathbb{E}_{p(\bm{x}, \bm{x}')}\left[\frac{1}{2}\left(\mathcal{L}_{\mathrm{CD}}(\bm{x}, \bm{x}')+\mathcal{L}_{\mathrm{CD}}(\bm{x}', \bm{x})\right)\right],
\end{equation}
where 
\begin{equation*}
\mathcal{L}_{\mathrm{CD}}(\bm{x}, \bm{x}') = \left(\pi_{+}-c(\bm{x}, \bm{x}')\right)\ell\left(g(\bm{x}),+1\right)+(\pi_{-}-c\left(\bm{x}, \bm{x}')\right)\ell\left(g(\bm{x}'),-1\right). 
\end{equation*}
\end{theorem}
Accordingly, we can derive an unbiased risk estimator for ConfDiff classification: 
\begin{align}\label{eq:confdiff_ure}
\widehat{R}_{\rm CD}(g) = \frac{1}{2n}\sum\nolimits_{i=1}^{n}\left(\mathcal{L}_{\mathrm{CD}}\left(\bm{x}_{i}^{\mathrm{CD}}, \bm{x}_{i}^{\mathrm{CD}'}\right)+\mathcal{L}_{\mathrm{CD}}\left(\bm{x}_{i}^{\mathrm{CD}'}, \bm{x}_{i}^{\mathrm{CD}}\right)\right).
\end{align}

Actually, Eq.~(\ref{eq:confdiff_ure}) is one of the candidates of the unbiased risk estimator. We introduce the following lemma:
\begin{lemma}\label{lemma_mvre}
The following expression is also an unbiased risk estimator:
\begin{equation}\label{minimum_var}
\frac{1}{n}\sum\nolimits_{i=1}^{n}\left(\alpha \mathcal{L}_{\mathrm{CD}}\left(\bm{x}_{i}^{\mathrm{CD}}, \bm{x}_{i}^{\mathrm{CD}'}\right)+(1-\alpha)\mathcal{L}_{\mathrm{CD}}\left(\bm{x}_{i}^{\mathrm{CD}'}, \bm{x}_{i}^{\mathrm{CD}}\right)\right),
\end{equation}
where $\alpha\in[0,1]$ is an arbitrary weight.
\end{lemma}
Then, we introduce the following theorem:
\begin{theorem}\label{min_var_thm}
The unbiased risk estimator in Eq.~(\ref{eq:confdiff_ure}) has the minimum variance among all the candidate unbiased risk estimators in the form of Eq.~(\ref{minimum_var}) w.r.t.~$\alpha\in[0,1]$.
\end{theorem}
Theorem~\ref{min_var_thm} indicates the variance minimality of the proposed unbiased risk estimator in Eq.~(\ref{eq:confdiff_ure}), and we adopt this risk estimator in the following sections.
\subsubsection{Estimation Error Bound}\label{sec:eeb}
In this subsection, we elaborate the convergence property of the proposed risk estimator $\widehat{R}_{\rm CD}(g)$ by giving an estimation error bound. Let $\mathcal{G}=\left\{g:\mathcal{X}\mapsto \mathbb{R}\right\}$ denote the model class. It is assumed that there exists some constant $C_{\mathcal{G}}$ such that $\sup_{g\in\mathcal{G}}\|g\|_{\infty} \leq C_{\mathcal{G}}$ and some constant $C_{\ell}$ such that $\sup_{|z|\leq C_{\mathcal{G}}}\ell(z, y) \leq C_{\ell}$. We also assume that the binary loss function $\ell(z, y)$ is Lipschitz continuous for $z$ with a Lipschitz constant $L_{\ell}$. Let $g^{*}=\mathop{\arg\min}_{g\in\mathcal{G}}R^{\mathrm B}(g)$ denote the minimizer of the classification risk in Eq.~(\ref{eq:binary_class_risk}) and $\widehat{g}_{\rm CD}=\mathop{\arg\min}_{g\in\mathcal{G}}\widehat{R}_{\rm CD}(g)$ denote the minimizer of the unbiased risk estimator in Eq.~(\ref{eq:confdiff_ure}). The following theorem can be derived:
\begin{theorem}\label{eeb_confdiff}
For any $\delta > 0$, the following inequality holds with probability at least $1 - \delta$:
\begin{equation}
R^{\mathrm B}\left(\widehat{g}_{\rm CD}\right) - R^{\mathrm B}\left(g^{*}\right) \leq 8L_{\ell}\mathfrak{R}_{n}(\mathcal{G})+4C_{\ell}\sqrt{\frac{\ln 2/\delta}{2n}},
\end{equation}
where $\mathfrak{R}_{n}\left(\mathcal{G}\right)$ denotes the Rademacher complexity of $\mathcal{G}$ for unlabeled data with size $n$.
\end{theorem}
From Theorem~\ref{eeb_confdiff}, we can observe that as $n\rightarrow \infty$, $R^{\mathrm B}\left(\widehat{g}_{\rm CD}\right) \rightarrow R^{\mathrm B}\left(g^{*}\right)$ because $\mathcal{R}_{n}\left(\mathcal{G}\right) \rightarrow 0$ for all parametric models with a bounded norm, such as deep neural networks trained with weight decay~\citep{golowich2018size}. Furthermore, the estimation error bound converges in $\mathcal{O}_{p}\left(1/\sqrt{n}\right)$, where $\mathcal{O}_{p}$ denotes the order in probability, which is the optimal parametric rate for empirical risk minimization without making additional assumptions~\citep{mendelson2008lower}.
\subsubsection{Robustness of Risk Estimator}\label{sec:confdiff_robustness}
In the previous subsections, it was assumed that the class prior probability is known in advance. In addition, it was assumed that the ground-truth confidence difference of each unlabeled data pair is accessible. However, these assumptions can rarely be satisfied in real-world scenarios, since the collection of confidence difference is inevitably injected with noise. In this subsection, we theoretically analyze the influence of an inaccurate class prior probability and noisy confidence difference on the learning procedure. 

Let $\bar{\mathcal{D}}^{\mathrm{CD}}=\left\{((\bm{x}_{i}^{\mathrm{CD}}, \bm{x}_{i}^{\mathrm{CD}'}), \bar{c}_{i})\right\}_{i=1}^{n}$ denote $n$ unlabeled data pairs with noisy confidence difference, where $\bar{c}_{i}$ is generated by corrupting the ground-truth confidence difference $c_{i}$ with noise. Besides, let $\bar{\pi}_{+}$ denote the inaccurate class prior probability accessible to the learning algorithm. Furthermore, let $\bar{R}_{\rm CD}(g)$ denote the empirical risk calculated based on the inaccurate class prior probability and noisy confidence difference. Let $\bar{g}_{\rm CD}=\mathop{\arg\min}_{g\in\mathcal{G}}\bar{R}_{\rm CD}(g)$ denote the minimizer of $\bar{R}_{\rm CD}(g)$. Then the following theorem gives an estimation error bound: 
\begin{theorem}\label{noisy_eeb_confdiff}
Based on the assumptions of Theorem~\ref{eeb_confdiff}, for any $\delta > 0$, the following inequality holds with probability at least $1 - \delta$:
\begin{align}
R^{\mathrm B}\left(\bar{g}_{\rm CD}\right) - R^{\mathrm B}\left(g^{*}\right) \leq &16L_{\ell}\mathfrak{R}_{n}\left(\mathcal{G}\right)+8C_{\ell}\sqrt{\frac{\ln{2/\delta}}{2n}}\nonumber\\
&+\frac{4C_{\ell}\sum\nolimits_{i=1}^{n}\left|\bar{c}_{i}-c_{i}\right|}{n}+4C_{\ell}\left|\bar{\pi}_{+}-\pi_{+}\right|.
\end{align}
\end{theorem}
Theorem~\ref{noisy_eeb_confdiff} indicates that the estimation error is bounded by twice the original bound in Theorem~\ref{eeb_confdiff} with the mean absolute error of the noisy confidence difference and the inaccurate class prior probability. Furthermore, if $\sum\nolimits_{i=1}^{n}\left|\bar{c}_{i}-c_{i}\right|$ has a sublinear growth rate with high probability and the class prior probability is estimated consistently, the risk estimator can be even consistent. This elaborates the robustness of the proposed approach. 
\subsubsection{Risk-Correction Approach}\label{sec:confdiff_cre}
It is worth noting that the empirical risk in Eq.~(\ref{eq:confdiff_ure}) may be negative due to negative terms, which is unreasonable because of the non-negative property of loss functions. This phenomenon will result in severe overfitting problems when complex models are adopted~\citep{lu2020mitigating,cao2021learning,feng2021pointwise}. To circumvent this difficulty, we wrap the individual loss terms in Eq.~(\ref{eq:confdiff_ure}) with \emph{risk-correction functions} proposed in~\citet{lu2020mitigating}, such as the rectified linear unit (ReLU) function $h(z) = \max(0, z)$ and the absolute value function $h(z) = |z|$. In this way, the corrected risk estimator for ConfDiff classification can be expressed as follows:
\begin{align} \label{corrected_ure}
\widetilde{R}_{\rm CD}(g) = &\frac{1}{2n}h\left(\sum\nolimits_{i=1}^{n}\left(\pi_{+}-c_{i}\right)\ell\left(g\left(\bm{x}_{i}^{\mathrm{CD}}\right),+1\right)\right) \nonumber \\
&+\frac{1}{2n}h\left(\sum\nolimits_{i=1}^{n}\left(\pi_{-}-c_{i}\right)\ell\left(g\left(\bm{x}_{i}^{\mathrm{CD}'}\right),-1\right)\right) \nonumber\\
&+\frac{1}{2n}h\left(\sum\nolimits_{i=1}^{n}\left(\pi_{+}+c_{i}\right)\ell\left(g\left(\bm{x}_{i}^{\mathrm{CD}'}\right),+1\right)\right)\nonumber\\
&+\frac{1}{2n}h\left(\sum\nolimits_{i=1}^{n}\left(\pi_{-}+c_{i}\right)\ell\left(g\left(\bm{x}_{i}^{\mathrm{CD}}\right),-1\right)\right).
\end{align}

We assume that the risk-correction function $h(z)$ is Lipschitz continuous with Lipschitz constant $L_{h}$. For ease of notation, we introduce
\begin{align}
\widehat{A}({g})&=\frac{1}{2n}\sum\nolimits_{i=1}^{n}(\pi_{+}-c_{i})\ell\left(g\left(\bm{x}_{i}^{\mathrm{CD}}\right),+1\right), \nonumber\\
\widehat{B}({g})&=\frac{1}{2n}\sum\nolimits_{i=1}^{n}(\pi_{-}-c_{i})\ell\left(g\left(\bm{x}_{i}^{\mathrm{CD}'}\right),-1\right),\nonumber\\
\widehat{C}({g})&=\frac{1}{2n}\sum\nolimits_{i=1}^{n}(\pi_{+}+c_{i})\ell\left(g\left(\bm{x}_{i}^{\mathrm{CD}'}\right),+1\right),\nonumber\\
\widehat{D}({g})&=\frac{1}{2n}\sum\nolimits_{i=1}^{n}(\pi_{-}+c_{i})\ell\left(g\left(\bm{x}_{i}^{\mathrm{CD}}\right),-1\right).
\end{align}
We assume that there exist positive constants $a,b,c,$ and $d$ such that $\mathbb{E}\left[\widehat{A}({g})\right] \geq a,\mathbb{E}\left[\widehat{B}({g})\right] \geq b,\mathbb{E}\left[\widehat{C}({g})\right] \geq c, $ and $\mathbb{E}\left[\widehat{D}({g})\right] \geq d$. Besides, let $\widetilde{g}_{\rm CD}=\mathop{\arg\min}_{g\in\mathcal{G}}\widetilde{R}_{\rm CD}(g)$ denote the minimizer of $\widetilde{R}_{\rm CD}(g)$. Then, Theorem~\ref{rc_consist} is provided to elaborate the bias and consistency of $\widetilde{R}_{\rm CD}(g)$.
\begin{theorem}\label{rc_consist}
Based on the assumptions of Theorem~\ref{eeb_confdiff}, the bias of the risk estimator $\widetilde{R}_{\rm CD}(g)$ decays exponentially as $n\rightarrow \infty$:
\begin{equation}
0\leq \mathbb{E}\left[\widetilde{R}_{\rm CD}(g)\right]-R^{\mathrm B}(g) \leq 2\left(L_{h}+1\right)C_{\ell}\Delta, 
\end{equation}
where $\Delta=\exp{\left(-2a^{2}n/C_{\ell}^{2}\right)}+\exp{\left(-2b^{2}n/C_{\ell}^{2}\right)}+\exp{\left(-2c^{2}n/C_{\ell}^{2}\right)}+\exp{\left(-2d^{2}n/C_{\ell}^{2}\right)}$. Furthermore, with probability at least $1-\delta$, we have
\begin{equation}
|\widetilde{R}_{\rm CD}(g)-R^{\mathrm B}(g)|\leq 2C_{\ell}L_{h}\sqrt{\frac{\ln{2/\delta}}{2n}}+2\left(L_{h}+1\right)C_{\ell}\Delta. 
\end{equation}
\end{theorem}
Theorem~\ref{rc_consist} demonstrates that $\widetilde{R}_{\rm CD}(g) \rightarrow R^{\mathrm B}(g)$ in $\mathcal{O}_{p}(1/\sqrt{n})$, which means that $\widetilde{R}_{\rm CD}(g)$ is biased yet consistent. The estimation error bound of $\widetilde{g}_{\rm CD}$ is analyzed in Theorem~\ref{rc_eeb_confdiff}.
\begin{theorem}\label{rc_eeb_confdiff}
Based on the assumptions of Theorem~\ref{rc_consist}, for any $\delta > 0$, the following inequality holds with probability at least $1 - \delta$:
\begin{equation}
R^{\mathrm B}\left(\widetilde{g}_{\rm CD}\right)-R^{\mathrm B}\left(g^{*}\right)\leq 8L_{\ell}\mathfrak{R}_{n}(\mathcal{G})+4C_{\ell}(L_{h}+1)\sqrt{\frac{\ln{2/\delta}}{2n}} + 4(L_{h}+1)C_{\ell}\Delta.
\end{equation}
\end{theorem}
Theorem~\ref{rc_eeb_confdiff} elucidates that as $n\rightarrow \infty$, $R^{\mathrm B}(\widetilde{g}_{\rm CD}) \rightarrow R^{\mathrm B}(g^{*})$, since $\mathcal{R}_{n}(\mathcal{G}) \rightarrow 0$ for all parametric models with a bounded norm~\citep{mohri2012foundations} and $\Delta \rightarrow 0$. Furthermore, the estimation error bound converges in $\mathcal{O}_{p}(1/\sqrt{n})$, which is the optimal parametric rate for empirical risk minimization without additional assumptions~\citep{mendelson2008lower}.

\section{Selected-Completely-at-Random Complementary-Label Learning}
\label{sec:scarce}
CL learning is a weakly supervised learning problem in which each training example is associated with one or multiple complementary labels indicating the classes to which it does not belong. Existing consistent approaches have relied on the uniform distribution assumption to model the generation of complementary labels, or on an ordinary-label training set to estimate the transition matrix in non-uniform cases. However, either condition may not be satisfied in real-world scenarios. In this section, we propose a novel consistent approach that does not rely on these conditions. Inspired by the positive-unlabeled~(PU) learning literature, we propose an \emph{unbiased risk estimator} based on the \emph{Selected-Completely-at-Random~(SCAR) assumption} for CL learning. We then introduce a \emph{risk-correction approach} to address overfitting problems. 
\subsection{Motivation}
CL learning is a weakly supervised learning problem that has received a lot of attention recently~\citep{ishida2017learning,feng2020learning,gao2021discriminative,liu2023consistent}. 
In CL learning, we are given training data associated with complementary labels that specify the classes to which the examples do not belong. The task is to learn a multi-class classifier that assigns correct labels to test data as in the standard supervised learning. Collecting training data with complementary labels is much easier and cheaper than collecting ordinary-label data. For example, when asking workers on crowdsourcing platforms to annotate training data, we only need to randomly select a candidate label and then ask them whether the example belongs to that class or not. Such ``yes'' or ``no'' questions are much easier to answer than asking workers to determine the ground-truth label from a large set of candidate labels. The benefits and effectiveness of CL learning have also been demonstrated in several machine learning problems and applications, such as domain adaptation~\citep{zhang2021learning,han2023rethinking}, semi-supervised learning~\citep{chen2020negative,ma2023rethinking,deng2024boosting},
noisy-label learning~\citep{kim2019nlnl}, adversarial robustness~\citep{zhou2022adversarial}, few-shot learning~\citep{wei2022an}, and medical image analysis~\citep{rezaei2020recurrent}.

Existing research works with \emph{consistency guarantees} have attempted to solve CL learning problems by making assumptions about the distribution of complementary labels. The remedy started with~\citet{ishida2017learning}, which proposed the \emph{uniform distribution assumption} that a label other than the ground-truth label is sampled from the uniform distribution to be the complementary label. A subsequent work extended it to arbitrary loss functions and models~\citep{ishida2019complementary} based on the same distribution assumption. Then,~\citet{feng2020learning} extended the problem setting to the existence of multiple complementary labels. Recent works have proposed discriminative methods that work by modeling the posterior probabilities of complementary labels instead of the generation process~\citep{chou2020unbiased,gao2021discriminative,liu2023consistent,lin2023reduction}. However, the uniform distribution assumption is still necessary to ensure the classifier consistency property~\citep{liu2023consistent}.~\citet{yu2018learning} proposed the \emph{biased distribution assumption}, elaborating that the generation of complementary labels follows a \emph{transition matrix}, i.e., the CL distribution is determined by the true label. 

In summary, previous CL learning approaches all require either the uniform distribution assumption or the biased distribution assumption to guarantee the consistency property, to the best of our knowledge.  However, such assumptions may not be satisfied in real-world scenarios. On the one hand, the uniform distribution assumption is too strong, since the transition probability for different complementary labels is undifferentiated, i.e., the transition probability from the true label to a complementary label is constant for all labels. Such an assumption is not realistic since the annotations may be imbalanced and biased~\citep{wei2023class,wang2025climage}. On the other hand, although the biased distribution assumption is more practical, an ordinary-label training set with \emph{deterministic labels}, also known as \emph{anchor points}~\citep{liu2015classification}, is essential for estimating transition probabilities during the training phase~\citep{yu2018learning}. However, the collection of ordinary-label data with deterministic labels is often unrealistic in CL learning problems~\citep{feng2020learning,gao2021discriminative}. Therefore, we are motivated to propose consistent methods that do not rely on previous strong assumptions.

\subsection{Data Generation Process}
Inspired by the SCAR assumption in PU learning~\citep{elkan2008learning,coudray2023risk}, we introduce the SCAR assumption for generating complementary labels, which can be summarized as follows.
\begin{assumption}[Selected-Completely-at-Random Assumption]\label{scar}
The CL data with the $k$-th class as a complementary label are sampled completely at random from the marginal density of the data not belonging to the $k$-th class, i.e.,
\begin{align}
p\left(k\in\bar{Y}|\bm{x},k\in{\mathcal{Y}\backslash \{y\}}\right)=p\left(k\in\bar{Y}|k\in{\mathcal{Y}\backslash \{y\}}\right)=c_{k},
\end{align}
where $c_{k}=\bar{\pi}_{k}/(1-\pi_k)$ is a constant specifying the fraction of data with the $k$-th class as a complementary label and $(\bm{x},y)$ is sampled from the density $p(\bm{x},y)$.
\end{assumption}

Our motivation is that complementary labels are often generated in a \emph{class-wise} manner. They can be collected by answering ``yes'' or ``no'' questions given a pair of an example and a candidate label~\citep{hu2019active,wang2021learning}. During an annotation round, we randomly select a candidate label and ask the annotators whether the example belongs to that class or not. The process is repeated iteratively, so that each example may be annotated with \emph{multiple} complementary labels. The SCAR assumption differs from the biased distribution assumption, where only one \emph{single} complementary label is generated by sampling only once from a multinomial distribution. Moreover, the SCAR assumption can be generalized to non-uniform cases by setting $c_k$ to different values for different labels. Therefore, our assumption is more practical in real-world scenarios. 

\subsection{Methodology}
In this section, we propose an unbiased risk estimator with the OVR strategy, followed by its theoretical analysis. Finally, we present a risk-correction approach to improve the generalization performance. Due to page limitations, detailed proofs of theorems and lemmas, as well as experimental results, can be found in \citet{wang2024learning}.

\subsubsection{OVR Strategy}\label{sec:scarce_ure}
The OVR strategy decomposes multi-class classification into a series of binary classification problems, which is a common strategy with extensive theoretical guarantees and sound performance~\citep{rifkin2004defense,zhang2004statistical}. Specifically, when considering a given class, examples that belong to the given class can be considered positive, while examples that do not belong to the given class can be considered negative. This naturally creates multiple binary classification problems. The OVR strategy instantiates the loss function $\mathcal{L}$ in Eq.~(\ref{eq:multi_class_risk}) with the OVR loss, i.e. 

\begin{equation}\label{eq:multi_class_risk_ovr}
R\left(f_{1}, f_{2}, \ldots, f_{q}\right)= \mathbb{E}_{p(\bm{x},y)}\left[\ell^{\mathrm{M}}\left(f_{y}\left(\bm{x}\right)\right)+\sum_{k\in{\mathcal{Y}\backslash \{y\}}}\ell^{\mathrm{M}}\left(-f_{k}\left(\bm{x}\right)\right)\right].
\end{equation}

Here, $f_{k}$ is a binary classifier w.r.t.~the $k$-th class, $\mathbb{E}$ denotes the expectation, and $\ell^{\mathrm{M}}:\mathbb{R}\rightarrow \mathbb{R}_{+}$ is a non-negative binary-class loss function.
Then, the predicted label for a test instance $\bm{x}$ is determined as 

\begin{equation}    
f(\bm{x}) = \mathop{\arg\max}_{k\in\mathcal{Y}}~f_{k}(\bm{x}).
\end{equation}

The goal is to find optimal classifiers $f_{1}^{*},f_{2}^{*},\ldots,f_{q}^{*}$ in a function class $\mathcal{F}$ which achieve the minimum classification risk in Eq.~(\ref{eq:multi_class_risk_ovr}), i.e., 
\begin{equation}
\left(f_{1}^{*},f_{2}^{*},\ldots,f_{q}^{*}\right) = \mathop{\arg\min}_{f_{1}, f_{2}, \ldots, f_{q}\in\mathcal{F}}~R\left(f_{1}, f_{2}, \ldots, f_{q}\right). 
\end{equation}
We show that the OVR risk can be rewritten using densities $p\left(\bm{x}|\bar{y}_{k}=1\right)$ and $p\left(\bm{x}|\bar{y}_{k}=0\right)$ as well. 
\begin{theorem}\label{ure}
When the OVR loss is used, the classification risk in Eq.~(\ref{eq:multi_class_risk_ovr}) can be equivalently expressed as $R(f_{1}, f_{2}, \ldots, f_{q})=\sum_{k=1}^{q}R_{k}(f_{k})$, where
\begin{align}
R_{k}(f_k) = &\mathbb{E}_{p\left(\bm{x}|\bar{y}_{k}=1\right)}\left[(1-\pi_{k})\ell^{\mathrm{M}}\left(-f_{k}(\bm{x})\right)+\left(\bar{\pi}_{k}+\pi_{k}-1\right)\right. \nonumber \\
&\left.\ell^{\mathrm{M}}\left(f_{k}(\bm{x})\right)\right]+\mathbb{E}_{p\left(\bm{x}|\bar{y}_{k}=0\right)}\left[\left(1-\bar{\pi}_{k}\right)\ell^{\mathrm{M}}\left(f_{k}(\bm{x})\right)\right]. 
\end{align}
\end{theorem}
Since the true densities $p\left(\bm{x}|\bar{y}_{k}=1\right)$ and $p\left(\bm{x}|\bar{y}_{k}=0\right)$ are not directly accessible, we approximate the risk \emph{empirically}. Suppose we have binary-class datasets $\mathcal{D}^{\rm N}_{k}$ and $\mathcal{D}^{\rm U}_{k}$ sampled i.i.d.~from $p\left(\bm{x}|\bar{y}_{k}=1\right)$ and $p\left(\bm{x}|\bar{y}_{k}=0\right)$, respectively. Then, an unbiased risk estimator can be derived from these binary-class datasets to approximate the classification risk in Theorem~\ref{ure} as $\widehat{R}(f_{1}, f_{2}, \ldots, f_{q})=\sum_{k=1}^{q}\widehat{R}_{k}(f_k)$, where 
\begin{align}\label{ure_eq}
\widehat{R}_{k}(f_k)=&\frac{1}{n^{\rm N}_{k}}\sum_{i=1}^{n^{\rm N}_{k}}\left(\left(1-\pi_{k}\right)\ell^{\mathrm{M}}\left(-f_{k}(\bm{x}_{k,i}^{\mathrm{N}})\right)+\left(\bar{\pi}_{k}+\pi_{k}-1\right)\right.\nonumber \\
&\left.\ell^{\mathrm{M}}\left(f_{k}(\bm{x}_{k,i}^{\mathrm{N}})\right)\right) +\frac{(1-\bar{\pi}_{k})}{n^{\rm U}_{k}}\sum_{i=1}^{n^{\rm U}_{k}}\ell^{\mathrm{M}}\left(f_{k}(\bm{x}_{k,i}^{\mathrm{U}})\right).
\end{align}
We may add regularization terms to $\widehat{R}\left(f_{1}, f_{2}, \ldots, f_{q}\right)$ when necessary~\citep{loshchilov2019decoupled}. This section considers generating the binary-class datasets $\mathcal{D}^{\rm N}_{k}$ and $\mathcal{D}^{\rm U}_{k}$ by \emph{duplicating} instances of $\mathcal{D}^{\mathrm{CL}}$. Specifically, if the $k$-th class is a complementary label of a training example, we regard its duplicated instance as a \emph{negative example} sampled from $p\left(\bm{x}|\bar{y}_{k}=1\right)$ and put the duplicated instance
in $\mathcal{D}^{\rm N}_{k}$. If the $k$-th class is not a complementary label of a training example, we regard its duplicated instance as an \emph{unlabeled example} sampled from $p\left(\bm{x}|\bar{y}_{k}=0\right)$ and put the duplicated instance
in $\mathcal{D}^{\rm U}_{k}$. In this way, we can obtain $q$ negative binary-class datasets and $q$ unlabeled binary-class datasets~($k\in \mathcal{Y}$):
\begin{align}
\mathcal{D}^{\rm N}_{k}&=\left\{(\bm{x}_{k,i}^{\mathrm{N}}, -1)\right\}_{i=1}^{n^{\rm N}_{k}}=\left\{(\bm{x}_{j}, -1)|(\bm{x}_{j},\bar{Y}_{j})\in\mathcal{D}^{\mathrm{CL}}, k\in\bar{Y}_{j}\right\};\label{neg_binary} \\
\mathcal{D}^{\rm U}_{k}&=\left\{\bm{x}_{k,i}^{\mathrm{U}}\right\}_{i=1}^{n^{\rm U}_{k}}=\left\{\bm{x}_{j}|(\bm{x}_{j},\bar{Y}_{j})\in\mathcal{D}^{\mathrm{CL}},k\notin\bar{Y}_{j}\right\}. \label{unlabel_binary}
\end{align}
When the class priors $\pi_{k}$ are not accessible to the learning algorithm, they can be estimated by off-the-shelf mixture proportion estimation approaches~\citep{scott2015rate,ramaswamy2016mixture,zhang2020unbiased,garg2021mixture,yao2022rethinking} with $\mathcal{D}^{\rm N}_{k}$ and $\mathcal{D}^{\rm U}_{k}$. Notably, the \emph{irreducibility}~\citep{blanchard2010semi,scott2013classification} assumption is necessary for class-prior estimation. However, it is still less demanding than the biased distribution assumption, which requires additional ordinary-label training data with deterministic labels, a.k.a.~anchor points, to estimate the transition matrix~\citep{yu2018learning}. 

\subsubsection{Theoretical Analysis}
In this section, we present theoretical analysis of the proposed method. 

We show that the proposed risk can be calibrated to the 0-1 loss \citep{zhang2004statistical}. Let $R_{\mathrm{0-1}}(f)=\mathbb{E}_{p(\bm{x},y)}\mathbb{I}(f(\bm{x})\neq y)$ denote the expected 0-1 loss where $f(\bm{x}) = \mathop{\arg\max}_{k\in\mathcal{Y}}~f_{k}(\bm{x})$ and $R_{\mathrm{0-1}}^{*}=\mathop{\min}_{f}~R_{\mathrm{0-1}}(f)$ denote the Bayes error. Besides, let $R^{*}= \mathop{\min}_{f_{1}, f_{2}, \ldots, f_{q}}~R(f_{1}, f_{2}, \ldots, f_{q})$ denote the minimum risk of the proposed risk. Then we have the following theorem.
\begin{theorem}\label{calibration}
Suppose the binary-class loss function $\ell^{\mathrm{M}}$ is convex, bounded below, differential, and satisfies $\ell^{\mathrm{M}}(z)\leq \ell^{\mathrm{M}}(-z)$ when $z > 0$. Then we have that for any $\epsilon_{1} > 0$, there exists an $\epsilon_{2} > 0$ such that
\begin{equation}
R\left(f_{1}, f_{2}, \ldots, f_{q}\right) \leq R^{*} + \epsilon_{2} \Rightarrow R_{\mathrm{0-1}}(f) \leq R_{\mathrm{0-1}}^{*} + \epsilon_{1}.
\end{equation}
\end{theorem}
The infinite-sample consistency elucidates that the proposed risk can be calibrated to the 0-1 loss. Therefore, if we minimize the proposed risk and obtain the optimal classifier, the classifier also achieves the Bayes error.

We further elaborate the convergence property of the empirical risk estimator $\widehat{R}(f_{1}, f_{2}, \ldots, f_{q})$ by providing its estimation error bound. The optimal classifiers w.r.t.~$\widehat{R}(f_{1}, f_{2}, \ldots, f_{q})$ are 
\begin{equation}
\left(\widehat{f}_{1}, \widehat{f}_{2}, \ldots, \widehat{f}_{q}\right) = \mathop{\arg\min}_{f_{1}, f_{2}, \ldots, f_{q}\in\mathcal{F}}~\widehat{R}\left(f_{1}, f_{2}, \ldots, f_{q}\right).
\end{equation}
We assume that there exists some constant $C_{\mathcal{F}}$ such that $\sup_{f_k\in\mathcal{F}}\|f_k\|_{\infty} \leq C_{\mathcal{F}}$ and some constant $C_{\ell^{\mathrm{M}}}$ such that $\sup_{|z|\leq C_{\mathcal{F}}}\ell^{\mathrm{M}}(z) \leq C_{\ell^{\mathrm{M}}}$. We also assume that the binary-class loss function $\ell^{\mathrm{M}}(z)$ is Lipschitz continuous w.r.t.~$z$ with a Lipschitz constant $L_{\ell^{\mathrm{M}}}$.
\begin{theorem}\label{eeb}
Based on the above assumptions, for any $\delta > 0$, the following inequality holds with probability at least $1 -\delta$:
\begin{align}
&R\left(\widehat{f}_{1}, \widehat{f}_{2}, \ldots, \widehat{f}_{q}\right) - R\left(f_{1}^{*},f_{2}^{*},\ldots,f_{q}^{*}\right)\leq \nonumber \\
&\sum_{k=1}^{q} \left(
(4-4\bar{\pi}_{k})L_{\ell^{\mathrm{M}}}\mathfrak{R}_{n^{\rm U}_{k},p^{\rm U}_{k}}(\mathcal{F})+(1-\bar{\pi}_{k})C_{\ell^{\mathrm{M}}}\sqrt{\frac{2\ln{\left(2/\delta\right)}}{n^{\rm U}_{k}}}\right. \nonumber \\
&\left.+(8-8\pi_{k}-4\bar{\pi}_{k})L_{\ell^{\mathrm{M}}}\mathfrak{R}_{n^{\rm N}_{k},p^{\rm N}_{k}}(\mathcal{F})+(2-2\pi_{k}-\bar{\pi}_{k})C_{\ell^{\mathrm{M}}}\sqrt{\frac{2\ln{\left(2/\delta\right)}}{n^{\rm N}_{k}}}\right),
\end{align}
where $\mathfrak{R}_{n^{\rm U}_{k},p^{\rm U}_{k}}(\mathcal{F})$ and $\mathfrak{R}_{n^{\rm N}_{k},p^{\rm N}_{k}}(\mathcal{F})$ denote the Rademacher complexity of $\mathcal{F}$ given $n^{\rm U}_{k}$ unlabeled data sampled from $p\left(\bm{x}|\bar{y}_{k}=0\right)$ and $n^{\rm N}_{k}$ negative data sampled from $p\left(\bm{x}|\bar{y}_{k}=1\right)$ respectively.
\end{theorem}
Theorem~\ref{eeb} elucidates an estimation error bound of our proposed risk estimator. When $n^{\rm U}_{k}$ and $ n^{\rm N}_{k} \rightarrow \infty$, $R\left(\widehat{f}_{1}, \widehat{f}_{2}, \ldots, \widehat{f}_{q}\right) \rightarrow R\left(f_{1}^{*},f_{2}^{*},\ldots,f_{q}^{*}\right)$ because $\mathfrak{R}_{n^{\rm U}_{k},p^{\rm U}_{k}}(\mathcal{F}) \rightarrow 0$ and $ \mathfrak{R}_{n^{\rm N}_{k},p^{\rm N}_{k}}(\mathcal{F}) \rightarrow 0$ for all parametric models with a bounded norm such as deep neural networks with weight decay~\citep{golowich2018size}.

\subsubsection{Risk-Correction Approach}\label{sec:scarce_cre}
Although the unbiased risk estimator~(URE) has sound theoretical properties, we have found that it can encounter several overfitting problems when using complex models such as deep neural networks~\citep{kiryo2017positive,lu2020mitigating,cao2021learning}. Therefore, following~\citet{lu2020mitigating,wang2023binary}, we wrap each potentially negative term with a \emph{non-negative risk-correction function} $g(z)$, such as the absolute value function $g(z)=|z|$. For ease of notation, we introduce 
\begin{equation}
\widehat{R}^{\rm P}_{k}(f_{k})=\frac{\bar{\pi}_{k}+\pi_{k}-1}{n^{\rm N}_{k}}\sum_{i=1}^{n^{\rm N}_{k}}\ell^{\mathrm{M}}\left(f_{k}(\bm{x}_{k,i}^{\mathrm{N}})\right)+\frac{1-\bar{\pi}_{k}}{n^{\rm U}_{k}}\sum_{i=1}^{n^{\rm U}_{k}}\ell^{\mathrm{M}}\left(f_{k}(\bm{x}_{k,i}^{\mathrm{U}})\right).
\end{equation}
Then, the corrected risk estimator can be written as $\widetilde{R}\left(f_{1}, f_{2}, \ldots, f_{q}\right)=\sum_{k=1}^{q}\widetilde{R}_{k}(f_{k})$, where
\begin{equation} \label{corrected_eq}
\widetilde{R}_{k}(f_{k})=g\left(\widehat{R}^{\rm P}_{k}(f_{k})\right)+\frac{1-\pi_{k}}{n^{\rm N}_{k}}\sum_{i=1}^{n^{\rm N}_{k}}\ell^{\mathrm{M}}\left(-f_{k}(\bm{x}_{k,i}^{\mathrm{N}})\right).
\end{equation}
It is obvious that Eq.~(\ref{corrected_eq}) is an upper bound of Eq.~(\ref{ure_eq}), so the bias is always present. Next, we perform a theoretical analysis to clarify that the corrected risk estimator is \emph{biased but consistent}. Since $\mathbb{E}\left[\widehat{R}^{\rm P}_{k}(f_{k})\right]=\pi_{k}\mathbb{E}_{p(\bm{x}|y=k)}\ell^{\mathrm{M}}\left(f_{k}(\bm{x})\right)$, we assume that there exists a \emph{positive constant} $\beta$ such that for $\forall k\in\mathcal{Y}, \mathbb{E}\left[\widehat{R}^{\rm P}_{k}(f_{k})\right] \geq \beta$. We also assume that the risk-correction function $g(z)$ is Lipschitz continuous with a Lipschitz constant $L_{g}$. Besides, we assume that there exists some constant $C_{\mathfrak{R}}$ such that the Rademacher complexity $\mathfrak{R}_{n,p}(\mathcal{F})$ for unlabeled~(with $n=n^{\rm U}_{k},p=p^{\rm U}_{k}$) and negative data~(with $n=n^{\rm N}_{k},p=p^{\rm N}_{k}$) satisfies $\mathfrak{R}_{n,p}(\mathcal{F}) \leq C_{\mathfrak{R}}/\sqrt{n}$. This assumption holds for many models, such as fully connected neural networks and linear-in-parameter models with a bounded norm~\citep{golowich2018size,lu2020mitigating}. 
We introduce $\left(\widetilde{f}_{1}, \widetilde{f}_{2}, \ldots, \widetilde{f}_{q}\right) = \mathop{\arg\min}_{f_{1}, f_{2}, \ldots, f_{q}\in\mathcal{F}}~\widetilde{R}\left(f_{1}, f_{2}, \ldots, f_{q}\right)$ and $\Delta_{k}=\exp{\left(-2\beta^{2}/\left((1 - \pi_{k}-\bar{\pi}_{k})^{2}C_{\ell^{\mathrm{M}}}^{2}/n^{\rm N}_{k}+(1-\bar{\pi}_{k})^{2}C_{\ell^{\mathrm{M}}}^{2}/n^{\rm U}_{k}\right)\right)}$. Then we have the following theorems.
\begin{theorem}\label{bias_and_consistency}
Based on the above assumptions, the bias of the expectation of the corrected risk estimator has the following lower and upper bounds: 
\begin{align}\label{bias_ineq}
0&\leq\mathbb{E}[\widetilde{R}(f_{1}, f_{2}, \ldots, f_{q})]-R(f_{1}, f_{2}, \ldots, f_{q}) \nonumber \\
&\leq \sum_{k=1}^{q}\left(2-2\bar{\pi}_{k}-\pi_{k}\right)\left(L_{g}+1\right)C_{\ell^{\mathrm{M}}}\Delta_{k}.
\end{align}
Furthermore, for any $\delta > 0$, the following inequality holds with probability at least $1-\delta$:
\begin{align}
&|\widetilde{R}(f_{1}, f_{2}, \ldots, f_{q})-R(f_{1}, f_{2}, \ldots, f_{q})| \nonumber \\
&\leq\mathcal{O}_{p}\left(\sum_{k=1}^{q}\left(1/\sqrt{n^{\rm N}_{k}}+1/\sqrt{n^{\rm U}_{k}}\right)\right).
\end{align}
\end{theorem}
\begin{theorem}\label{corrected_eeb}
Based on the above assumptions, for any $\delta>0$, the following inequality holds with probability at least $1-\delta$:
\begin{align}
&R(\widetilde{f}_{1}, \widetilde{f}_{2}, \ldots, \widetilde{f}_{q}) - R(f_{1}^{*},f_{2}^{*},\ldots,f_{q}^{*}) \nonumber \\
&\leq\mathcal{O}_{p}\left(\sum_{k=1}^{q}\left(1/\sqrt{n^{\rm N}_{k}}+1/\sqrt{n^{\rm U}_{k}}\right)\right). 
\end{align}
\end{theorem}
Theorem~\ref{bias_and_consistency} shows that $\widetilde{R}(f_{1}, f_{2}, \ldots, f_{q})\rightarrow R(f_{1}, f_{2}, \ldots, f_{q})$ as $n^{\rm U}_{k}$ and $ n^{\rm N}_{k} \rightarrow \infty$, indicating that the corrected risk estimator is biased but consistent. An estimation error bound is also shown in Theorem~\ref{corrected_eeb}. The convergence rate of the estimation error bound is still the same after employing the risk-correction function.

\section{Realistic Evaluation of Deep Partial-Label Learning Algorithms}
\label{sec:plench}
PL learning is a weakly supervised learning problem in which each example is associated with multiple candidate labels and only one is the true label. In recent years, many deep PL learning algorithms have been developed to improve model performance. However, we find that some early developed algorithms are often underestimated and can outperform many later algorithms with complicated designs. In this section, we delve into the empirical perspective of PL learning and identify several critical but previously overlooked issues. First, model selection for PL learning is non-trivial, but has never been systematically studied. Second, the experimental settings are highly inconsistent, making it difficult to evaluate the effectiveness of the algorithms. Based on these findings, we propose \textsc{Plench}, the first PL learning bENCHmark to systematically compare state-of-the-art deep PL learning algorithms. We investigate the model selection problem for PL learning for the first time, and propose novel model selection criteria with theoretical guarantees. Researchers can quickly and conveniently perform a comprehensive and fair evaluation and verify the effectiveness of newly developed algorithms based on \textsc{Plench}. We hope that \textsc{Plench} will facilitate standardized, fair, and practical evaluation of PL learning algorithms in the future.
\subsection{Motivation}\label{intro_sec}
PL learning is a weakly supervised learning problem that has attracted much attention recently~\citep{sugiyama2022machine,wang2022adaptive,tian2023partial}. In PL learning, each training example is associated with multiple candidate labels ~\citep{jin2002learning,cour2011learning}. The true label for each example is hidden in the set of candidate labels, but not accessible to the learning algorithm. PL learning has been successfully applied to computer vision~\citep{liu2012conditional,zeng2013learning,chen2018learning,gong2018regularization,tang2023disambiguated,wang2024learning}, natural language processing~\citep{garrette2013learning,zhou2018weakly,ren2016afet,ren2016label}, web mining~\citep{luo2010learning}, ecoinformatics~\citep{briggs2012rank,wang2019partial,li2021detecting,lyu2022self}, etc.

Among various strategies to address this problem, deep learning-based PL learning algorithms have demonstrated satisfactory generalization performance due to the strong representation learning capabilities of deep neural networks~\citep{lv2020progressive,wang2022pico}. Despite the abundance of algorithms in this area, we find that there are several fundamental and critical issues that have received less attention in the PL learning literature. First, most PL learning algorithms select their hyperparameters by using a clean ordinary-label validation set~\citep{qiao2023decompositional,xu2023alim,xu2023progressive}. However, the original definition of PL learning does not allow the existence of an ordinary-label dataset~\citep{jin2002learning,cour2011learning,zhang2017disambiguation}, indicating a mismatch between the problem definitions and experimental settings in the literature. This problem can even lead to \emph{unfair comparisons} if some algorithms follow the classical protocol of PL learning to prohibit the use of ordinary-label data, while some algorithms do not. Moreover, \emph{if we have a clean ordinary-label dataset, why not use it for training?} In many cases, the use of clean labels is more valuable for weakly supervised learning than for ordinary supervised learning~\citep{hendrycks2018using,yu2023delving}. 

Second, we have found that the experimental settings used in different papers are often quite different, creating a dilemma when comparing the performance of different algorithms. Such an obstacle can hinder objective comparisons of different algorithms, making it difficult to determine the effectiveness of a developed technique. These problems therefore motivate us to propose a unified evaluation framework for PL learning.

\subsection{Model Selection for PL Learning}\label{model_selection_section}
We follow the original definition of PL learning to have only a single PL training set~\citep{cour2011learning,zhang2017disambiguation}. Then, following a widely used validation procedure in machine learning~\citep{raschka2018model,gulrajani2021in}, we divide a PL validation set $\mathcal{D}^{\rm Val}=\{\left(\bm{x}^{\rm Val}_i,S^{\rm Val}_i\right)\}_{i=1}^{n^{\rm Val}}$ from the training set for model selection. Next, we introduce each model selection criterion in turn. Due to page limits, detailed proofs of the theorems and propositions can be found in \citet{wang2025realistic}.
\subsubsection{Covering Rate}
\begin{definition}[Covering Rate~(CR)]The Covering Rate of a multi-class classifier $\bm{f}$ on the PL validation set $\mathcal{D}^{\rm Val}$ is defined as:
\begin{equation}
{\rm CR}(\bm{f})=\frac{1}{n^{\rm Val}}\sum_{i=1}^{n^{\rm Val}}\mathbb{I}\left(\mathop{\arg\max}_{j}~f_{j}(\bm{x}^{\rm Val}_i)\in S^{\rm Val}_i\right).
\end{equation}
\end{definition}
CR indicates the fraction of validation data whose predicted label is included in its candidate label set. It is a natural metric in PL learning, but its effectiveness may depend on the size of the candidate label sets. When the number of partial labels of each example is equal to 1, i.e., $|S|=1$, PL learning is reduced to ordinary supervised learning and CR is reduced to the validation accuracy. However, as $|S|$ increases, more false positive labels are included in the candidate label set. In the most extreme case, when $|S|=q$, PL learning is reduced to unsupervised learning and CR does not convey effective information. Before analyzing the gap between CR and the validation accuracy, we introduce the following definition.
\begin{definition}[Ambiguity Degree]\label{sadc}The Ambiguity Degree $\gamma$ is defined as 
\begin{equation}
\gamma = \sup_{(\bm{x},y)\sim p(\bm{x},y),S\sim p(S|\bm{x},y),\bar{y}\neq y} p(\bar{y}\in S),
\end{equation}
where $p(\bm{x},y)$ is the joint distribution over $\bm{x}$ and $y$.
\end{definition}
If $\gamma<1$, the small ambiguity degree is satisfied and the ERM learnability for PL learning is guaranteed~\citep{cour2011learning,liu2014learnability}. We also define the expected accuracy as 
\begin{equation}
{\rm ACC}\left(\bm{f}\right)=\mathbb{E}_{p(\bm{x},y)}\mathbb{I}(\mathop{\arg\max}_{l}~f_{l}\left(\bm{x}\right)= y).
\end{equation}
Then, we have the following proposition.
\begin{proposition}\label{cr_bound}
Suppose that there is a constant $\epsilon\in\left(0,1\right)$ such that the expected accuracy of a classifier $\bm{f}$ satisfies ${\rm ACC}\left(\bm{f}\right)\geq\epsilon$. Then, we have $\mathbb{E}\left[{\rm CR}(\bm{f})\right]-{\rm ACC}\left(\bm{f}\right)\leq(1-\epsilon)\gamma$.
\end{proposition}
The gap between CR and the accuracy is affected by both the accuracy and the ambiguity degree. If the classifier is more accurate and the PL dataset is less ambiguous, the gap between the CR metric and the accuracy will be smaller. Furthermore, we show that the minimizers of both are the same under certain conditions.
\begin{theorem}\label{cr_consistency}
Suppose that the partial labels are generated by following the USS or the FPS with a constant flipping probability. Then, for any two classifiers $\bm{f}_1$ and $\bm{f}_2$ that satisfy $\mathbb{E}\left[{\rm CR}(\bm{f}_1)\right] < \mathbb{E}\left[{\rm CR}(\bm{f}_2)\right]$, we have ${\rm ACC}\left(\bm{f}_1\right) < {\rm ACC}\left(\bm{f}_2\right)$.
\end{theorem}
Theorem~\ref{cr_consistency} shows that when partial labels are generated by using the USS or the FPS with a constant flipping probability, the classifier that minimizes the expectation of CR will also minimize the expected accuracy. Therefore, the CR metric will serve as a \emph{consistent} model selection criterion for PL learning under certain data distribution assumptions. However, this conclusion may not hold when partial labels are not generated by either strategy.

\subsubsection{Approximated Accuracy}
Next, we introduce the definition of the Approximated Accuracy metric.
\begin{definition}[Approximated Accuracy~(AA)]The Approximated Accuracy of a multi-class classifier $\bm{f}$ on the PL validation set $\mathcal{D}^{\rm Val}$ is defined as:
\begin{equation}
\mathrm{AA}(\bm{f})=\frac{1}{n^{\rm Val}}\sum_{i=1}^{n^{\rm Val}}\sum_{j\in S^{\rm Val}_i}\frac{f_{j}(\bm{x}^{\rm Val}_i)}{\sum_{k\in S^{\rm Val}_i}f_{k}(\bm{x}^{\rm Val}_i)}\mathbb{I}\left(\mathop{\arg\max}_{l}~f_{l}(\bm{x}^{\rm Val}_i)= j\right).
\end{equation}
\end{definition}
Then, we have the following theorem.
\begin{theorem}\label{ure_thm}
Suppose there is a function $C: \mathcal{X}\bigtimes2^{\mathcal{Y}}\mapsto\mathbb{R}$ such that the condition $p(S|\bm{x}, y)=C(\bm{x}, S)\mathbb{I}\left(y\in S\right)$ holds for PL data. Suppose further that the multi-class classifier $\bm{f}(\bm{x})$ is consistent with $p(y|\bm{x})$. Then, under mild conditions $\mathrm{AA}(\bm{f})$ is statistically consistent with the expected accuracy, i.e., $\mathbb{E}\left[{\rm AA}(\bm{f})\right]={\rm ACC}\left(\bm{f}\right)$.
\end{theorem}
The introduction of AA is inspired by data-generation-based strategies for PL learning~\citep{wu2023learning}. Theorem~\ref{ure_thm} illustrates that AA can be a consistent metric w.r.t.~the expected classification accuracy on test data under certain assumptions. In particular, the data distribution assumption can hold for a wide range of types of partial labels~\citep{liu2012conditional,wu2023learning}. However, there are two factors that may affect its effectiveness. First, it may not be suitable for certain algorithms~\citep{wen2021leveraged} where the loss function is not \emph{strictly proper}~\citep{gneiting2007strictly,charoenphakdee2021on} and its modeling output is not calibrated to the posterior probabilities. Second, it requires that the modeling output to be accurate, which may not be satisfied in the early stages of training. 
\subsubsection{Oracle Accuracy}
Finally, we present the definition of the Oracle Accuracy metric.
\begin{definition}[Oracle Accuracy~(OA)]
The Oracle Accuracy of a multi-class classifier $\bm{f}$ on a PL validation set $\mathcal{D}^{\rm Val}$ is defined as
\begin{equation}
\mathrm{OA}(\bm{f})=\frac{1}{n^{\rm Val}}\sum_{i=1}^{n^{\rm Val}}\mathbb{I}\left(\mathop{\arg\max}_{l}~f_{l}(\bm{x}^{\rm Val}_i)= y^{\rm Val}_i\right),
\end{equation}
where $y^{\rm Val}_i$ is the underlying true label of $\bm{x}^{\rm Val}_i$.
\end{definition}
The OA metric is natural if we have access to true labels in supervised learning. However, such a condition is not realistic in real problems of PL learning. In fact, the availability of true labels contradicts the original motivation of PL learning, which is to reduce labeling costs at the expense of true label ignorance. Inspired by~\citet{gulrajani2021in}, we restrict the use of true labels by allowing only one query (the last checkpoint) for each hyperparameter configuration. This means that we do not allow early stopping when using the OA metric. We try to compensate for the unrealistic access to true labels by restricting the model selection space. We also include the results of OA with early stopping (ES), which can be considered as an upper bound of the model selection performance, for reference. Actually, OA with ES may be the most common model selection criterion for PL learning papers.

\subsection{PLENCH}\label{sec:plench_settings}
In this section, we introduce the benchmark algorithms, datasets, and settings. Due to page limits, the detailed benchmark results can be found in \citet{wang2025realistic}.
\subsubsection{Benchmark Algorithms}
We have divided the benchmark algorithms into four main groups. 

We considered deep PL learning algorithms that used simple loss functions or disambiguation strategies without strong regularization techniques as vanilla deep PL learning algorithms. Identification-based strategies included PRODEN~\citep{lv2020progressive}, CAVL~\citep{zhang20222exploiting}, and POP~\citep{xu2023progressive}. Note that RC~\citep{feng2020provably} is in the same form as PRODEN, so we only included PRODEN here to avoid repetitive comparisons. The averaging-based strategies included ABS-MAE~\citep{lv2024on} and ABS-GCE~\citep{lv2024on}, which showed favorable performance among the losses in the family. 
The data-generation-based strategies included EXP~\citep{feng2020learning}, MCL-GCE~\citep{feng2020learning}, MCL-MSE~\citep{feng2020learning}, CC~\citep{feng2020provably}, LWS~\citep{wen2021leveraged}, and IDGP~\citep{qiao2023decompositional}. Also, LOG~\citep{feng2020learning} is analogous to CC, so we only included CC here.

CL learning is a special case of PL learning with only one label excluded from each candidate label set of training examples, i.e., $|S|=q-1$~\citep{wang2025climage}. If we consider each label outside the candidate label set as a complementary label, it is possible to apply CL learning algorithms to solve PL learning problems as well~\citep{feng2020learning}. Therefore, we included several vanilla CL learning algorithms with simple loss functions in \textsc{Plench}. The vanilla CL learning algorithms included PC~\citep{ishida2017learning}, Forward~\citep{yu2018learning}, NN~\citep{ishida2019complementary}, GA~\citep{ishida2019complementary}, SCL-EXP~\citep{chou2020unbiased}, SCL-NL~\citep{chou2020unbiased}, L-W~\citep{gao2021discriminative}, and OP-W~\citep{liu2023consistent}.

We considered PL learning algorithms that employ strong representation learning or regularization techniques to be holistic PL learning algorithms. We used five algorithms, including VALEN~\citep{xu2021instance}, PiCO~\citep{wang2022pico}, ABLE~\citep{xia2022ambiguity}, CRDPL learning~\citep{wu2022revisiting}, and DIRK~\citep{wu2024distilling}.

We also included three noisy PL learning algorithms that explicitly considered tailored strategies to handle the cases where the true label might be outside the candidate label set. The deep noisy PL learning algorithms included FREDIS~\citep{qiao2023fredis}, ALIM~\citep{xu2023alim}, and PiCO+~\citep{wang2024picoplus}.
\subsubsection{Benchmark Datasets}
To comprehensively evaluate the model performance of all algorithms, we included eleven real-world PL learning benchmark datasets, including nine widely used tabular datasets and two image datasets that we collected. The tabular datasets included Lost~\citep{cour2011learning}, Soccer Player~\citep{zeng2013learning}, and Yahoo! News~\citep{guillaumin2010multiple} for the automatic face naming task, MSRCv2~\citep{liu2012conditional} for the object classification task, Mirflickr~\citep{huiskes2008mir} for the web image classification task, Birdsong~\citep{briggs2012rank} for the bird song classification task, Malagasy~\citep{garrette2013learning}, Italian~\citep{johan2009converting}, and English~\citep{zhou2018weakly} for the POS tagging task.

For face age estimation, we considered ten crowdsourced labels of age numbers along with the true label as partial labels for a given human face. For automatic face naming, we considered the names in the corresponding captions or subtitles as partial labels for a face cropped from an image. For object classification, we considered object classes as partial labels for a segmentation part in an image. For web image classification, we considered tags on a web page as partial labels for a given image. For bird song classification, we considered bird species appearing in a ten-second bird song fragment as partial labels for the singing syllables of the fragment. For POS tagging, we considered all possible POS tags as partial labels for a given word with its contexts.

\subsubsection{Benchmark Settings}
In this section, we mainly investigate the hyperparameter tuning problem in the context of model selection. For tabular datasets, we first divided a test set from the entire dataset. Since the datasets were not explicitly divided into training and validation parts, we manually divided them into a PL training set $\mathcal{D}^{\rm Tr}$ and a PL validation set $\mathcal{D}^{\rm Val}$. Then, we trained a model with $\mathcal{D}^{\rm Tr}$ and evaluated its validation performance on $\mathcal{D}^{\rm Val}$ with the model selection criteria proposed in Section~\ref{model_selection_section} as well as its test performance on the test set $\mathcal{D}^{\rm Te}$ with ordinary labels. We then selected the checkpoint with the best validation performance and returned the corresponding test accuracy as the final result. We randomly selected a set of hyperparameter configurations from a given pool for a given data split and recorded the mean accuracy as well as the standard deviations for the best performance with different dataset splits. We used ResNet~\citep{he2016deep} and DenseNet~\citep{huang2017densely} for image datasets and a multilayer perceptron~(MLP) with a hidden layer width of 500 equipped with the ReLU~\citep{nair2010rectified} activation function for tabular datasets. 

All the algorithms were implemented in PyTorch~\citep{paszke2019pytorch} and all experiments were conducted with a single NVIDIA Tesla V100 GPU. We used the Adam optimizer~\citep{kingma2015adam}. We ran 60,000 iterations for the image datasets, 20,000 iterations for the Soccer Player, Italian, Yahoo! News, and English datasets, and 10,000 iterations for the other datasets. We recorded the performance on validation and test sets per 1,000 iterations. We used three random data splits for PLCIFAR10 and five random data splits for tabular datasets. For each data split, we selected 20 random hyperparameter configurations from a given pool.
 
\section{Conclusion}
In this chapter, we presented several recent advances in weakly supervised learning, including new supervision paradigms, assumption relaxations, and practical solutions. First, we delved into a novel weakly supervised learning setting in which only unlabeled data pairs with confidence difference were provided. To solve this problem, we derived an unbiased risk estimator to perform empirical risk minimization. We established an estimation error bound to demonstrate that the optimal parametric convergence rate can be attained. Additionally, we introduced a risk correction approach to address overfitting issues. Second, we made the first attempt at consistent complementary-label learning that does not rely on the uniform distribution assumption or an ordinary-label training set to estimate the transition matrix in non-uniform cases. Based on a more practical distribution assumption, we proposed a consistent approach with theoretical guarantees. Third, we proposed the first benchmark for performance evaluation of state-of-the-art deep PLL algorithms. We introduced new model selection criteria to address the lack of criteria for selecting models in PLL. In the future, it is promising to investigate the integration of foundation models~\citep{li2026weakly}, evaluation issues of other WSL problems~\citep{wang2026accessible}, and consistent approaches for more WSL problems~\citep{ma2026,wang2026rethinking}.
\section*{Acknowledgment}
The authors would like to thank the following individuals for their contributions to the conference papers included in this chapter: Lei Feng, Yuchen Jiang, Min-Ling Zhang, Takashi Ishida, Yu-Jie Zhang, Dong-Dong Wu, and Jindong Wang. WW was supported by the SGU MEXT Scholarship, by the Junior Research Associate (JRA) program of
RIKEN, and by Microsoft Research Asia. MS was supported by JST ASPIRE Grant Number JPMJAP2405.
\bibliographystyle{plainnat}
\bibliography{ref}

\end{document}